\pdfoutput=1

\documentclass[11pt]{article}

\usepackage[preprint]{acl}

\usepackage{times}
\usepackage{latexsym}

\usepackage[T1]{fontenc}

\usepackage[utf8]{inputenc}

\usepackage{microtype}

\usepackage{inconsolata}

\usepackage{graphicx}

\usepackage{hyperref}
\usepackage{url}
\usepackage{graphicx}
\usepackage{amsmath}
\usepackage{amsthm}
\usepackage{amssymb}
\usepackage{multirow}
\usepackage{tabularx}
\usepackage{setspace}
\usepackage{enumitem}
\usepackage{courier}
\usepackage{bm}
\usepackage{geometry}
\usepackage{algorithm}
\usepackage{algorithmicx}
\usepackage{algpseudocode}
\usepackage[algo2e]{algorithm2e}
\usepackage{makecell}
\usepackage{longtable}
\usepackage[utf8]{inputenc}
\usepackage{array}
\usepackage{listings}

\usepackage{booktabs} 
\usepackage{colortbl} 
\usepackage{xcolor}   
\usepackage{tcolorbox}                   
\usepackage{subcaption}

\definecolor{lightblue}{RGB}{173,216,230} 

\title{RankPO: Preference Optimization for Job-Talent Matching}

\author{
 \textbf{Yafei Zhang},
 \textbf{Murray Wang},
 \textbf{Yu Wang},
 \textbf{Xiaohui Wang}
\\
\\
 Laboratory for AI-Powered Financial Technologies, 
 \\ City University of Hong Kong, Hong Kong S.A.R., China
}

\begin{document}
\maketitle
\begin{abstract}

Matching job descriptions (JDs) with suitable talent requires models capable of understanding not only textual similarities between JDs and candidate resumes
but also contextual factors such as geographical location and academic seniority. 
To address this challenge, we propose a two-stage training framework for large language models (LLMs). 
In the first stage, a contrastive learning approach is used to train the model on a dataset constructed from real-world matching rules, such as geographical alignment and research area overlap. 
While effective, this model primarily learns patterns that 
defined by the matching rules.
In the second stage, we introduce a novel preference-based fine-tuning method inspired by 
Direct Preference Optimization (DPO), termed Rank Preference Optimization (RankPO), to align the model with AI-curated pairwise preferences emphasizing textual understanding. 
Our experiments show that while the first-stage model achieves strong performance on rule-based data (nDCG@20 = 0.706), it lacks robust textual understanding (alignment with AI annotations = 0.46). 
By fine-tuning with RankPO, we achieve a balanced model that retains relatively good performance in the original tasks while significantly improving the alignment with AI preferences.
The code and data are available at \url{https://github.com/yflyzhang/RankPO}.

\end{abstract}

\section{Introduction}

\begin{figure*}[t]
  \centering
  \includegraphics[width=0.9\linewidth]{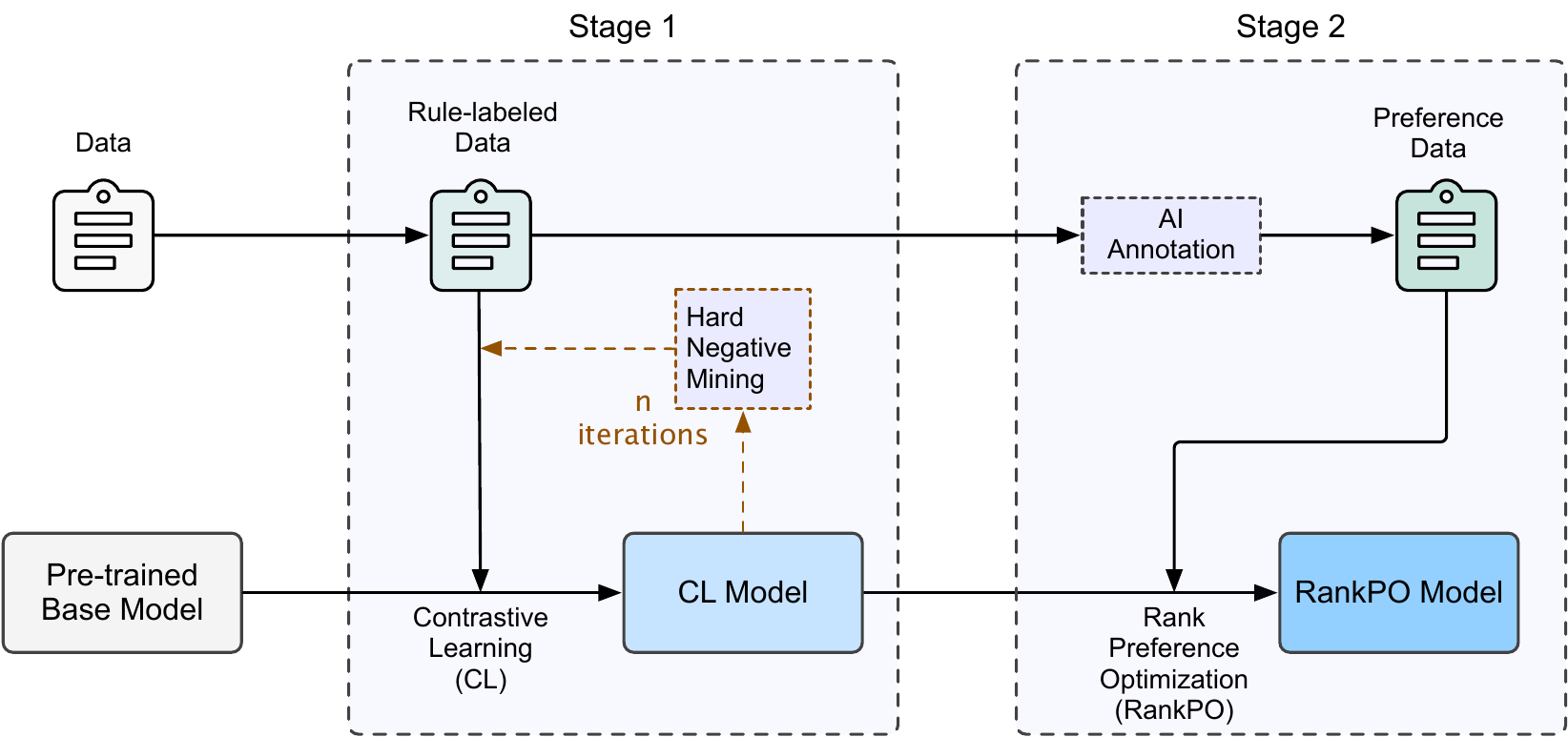}
  \caption {Illustration of the proposed two-stage training framework for job-talent matching. 
  The framework consists of two stages: (1) a contrastive learning stage to acquire foundational capabilities, and (2) a fine-tuning stage using RankPO to improve alignment with AI preferences while striving to retain capabilities learned from the first stage.
  }
  \label{fig:system}
\end{figure*}

Matching job descriptions (JDs) with suitable candidates is an important yet challenging task, especially in specialized domains such as academia~\citep{fredriksson2018mismatch}. Recruitment decisions often rely on nuanced factors such as research expertise, academic achievements, geographical preferences, and alignment with institutional goals. 
Traditional systems often rely on rule-based methods or textual similarity measures, which may fail to capture the nuanced requirements of JD-Talent matching.
Modern machine learning models, particularly large language models (LLMs), offer an opportunity to bridge this gap through context-aware analysis~\citep{wu2024survey,li2024large,bao2024large,ren2024representation}.
In this work, we propose a two-stage training framework that combines 
contrastive learning and preference optimization to build a robust and adaptable JD-Talent matching system.
Figure~\ref{fig:system} illustrates the proposed framework.

JD-Talent matching in academia is a particularly challenging task due to the lack of large-scale, high-quality datasets that comprehensively represent the matching process. Existing publicly available datasets tend to focus on general-purpose job markets, which may not reflect the unique requirements of academic recruitment. 
Recruiting for academic positions often involves strict requirements regarding geographic proximity, academic seniority, and research expertise. These constraints are difficult to encode using purely textual similarity metrics.
For instance, a professor with extensive experience in a specific field might be highly relevant to a JD based on textual similarity alone, but they would not be a suitable candidate for a postdoctoral position due to their academic seniority.

Regarding the data scarcity challenges,
we first compile a dataset of matched JD-Talent pairs by collecting extensive author profiles from academic literature databases (e.g., Web of Science) and job postings from recruitment platforms (e.g., Science Careers). 
Positive JD-Talent pairs were formed based on realistic matching rules (e.g., geographical proximity, academic experience, and research expertise) to ensure the relevance of the dataset. 

Contrastive learning has emerged as a powerful technique for learning robust representations by encouraging similarities between positive pairs while separating negative pairs in the embedding space. Originally introduced in the context of unsupervised or self-supervised learning~\citep{chen2020simple}, the supervised variant leverages labeled data to explicitly define positive and negative pairs ~\citep{khosla2020supervised, robinson2021contrastive, gao2021simcse}. In our framework, we use contrastive learning to train the model on rule-based JD-Talent pairs. 
While this step provides a strong starting point, it inherently biases the model towards the specific rules used to construct the dataset, limiting its generalization and semantic understanding.

Preference-based fine-tuning has recently gained attention for its ability to align models with human-like decision-making through reinforcement learning. Reinforcement Learning from Human Feedback (RLHF)~\citep{christiano2017deep, stiennon2020learning, ouyang2022training, bai2022training}, as demonstrated in training models like OpenAI’s ChatGPT, leverages pairwise preference data to fine-tune models towards behaviors or outputs that align with human preferences. This approach is particularly useful when the training data is insufficient to capture complex evaluation criteria, such as semantic relevance in JD-Talent matching. However, RLHF often requires substantial human annotation efforts and computational resources and is sensitive to reward model design, making it difficult to implement in certain domains~\citep{zheng2023secrets}.

Direct Preference Optimization (DPO)~\citep{rafailov2023direct} is a recently proposed alternative to RLHF that simplifies the preference alignment process by directly optimizing model outputs based on pairwise preferences. 
Unlike RLHF, 
DPO and its variants~\citep{wang2024beyond, meng2024simpo, amini2024direct} avoid
the complexities of reward modeling and instead uses a contrastive loss to align pairs of outputs with human or AI preferences. 
This approach has shown great success in tasks where the input $x$ (e.g., a query or context) and output $y$ (e.g., a response or completion) are modeled together to compute a joint probability distribution $p(y\mid x)$, such as conditional text generation tasks.

However, in many ranking or retrieval tasks, 
such as job-talent matching,
we often need to compare a single query against a large pool of candidates.
It is common to encode
the input (e.g., a query or a job description) and output (e.g., a candidate profile) 
independently into a shared embedding space.
This encoding paradigm enables efficient large-scale retrieval, as embeddings for outputs (e.g., candidate profiles) can be precomputed and stored offline, significantly reducing computation at query time.
To address this limitation, we propose Rank Preference Optimization (RankPO), a tailored extension of DPO. RankPO modifies the preference optimization process to operate on independently encoded representations of $x$ and $y$, introducing a new loss function that optimizes pairwise preferences directly in the embedding space. This adaptation ensures that RankPO is both scalable and efficient, making it well-suited for large-scale ranking tasks such as job-talent matching.

In practice, we can leverage advanced AI models e.g., GPT-4o~\citep{hurst2024gpt}, DeepSeek V3~\citep{liu2024deepseek}, to curate pairwise preference data emphasizing semantic relevance in JD-Talent matching. 
The model can then be optimized
with these curated preference data through the RankPO framework. 
This step ensures that the final model captures the deeper textual understanding exhibited by advanced AI systems while
striving to retain the capabilities obtained from the previous stage.

Our contributions are as follows:

1. We construct a large-scale JD-Talent dataset by collecting author profiles from academic literature databases and job descriptions from academic recruitment platforms. 
This dataset is enriched with positive samples generated under pre-defined matching rules.
While imperfect, this dataset serves as a valuable benchmark and 
provides a foundation for further research and improvements in this area.

2. We introduce RankPO, a preference optimization framework inspired by DPO.
This is a novel extension of DPO that operates on independently encoded input-output pairs.
RankPO provides a scalable and efficient solution for ranking tasks, making it broadly applicable beyond the JD-talent matching use case.

3. We present a unified framework that integrates contrastive learning and preference optimization. 
Using contrastive learning to capture meaningful relationships and RankPO to optimize preferences, our framework offers a robust and flexible solution to build ranking systems, especially in scenarios with limited labeled data.

\section{Dataset}

\subsection{Data Sources}
Our dataset consists of job postings and researcher profiles.
Academic job postings are collected from several major academic recruitment platforms such as Times Higher Education\footnote{\url{https://www.timeshighereducation.com/unijobs/listings/}}, Science Careers\footnote{\url{https://jobs.sciencecareers.org/}}, and Nature Careers\footnote{\url{https://www.nature.com/naturecareers/}}, capturing details such as required qualifications, research focus, and institutional priorities.
For researcher profiles,
we curate a large-scale academic dataset,
aggregated from academic databases such as Microsoft Academic Graph (MAG)~\citep{sinha2015anoverview}, Web of Science (WOS), Google Scholar, and ORCID. 
The original dataset covers more than 200 million academic papers and associated author information, including publication records, research areas, and affiliations.

\subsection{Data Processing}

Using pre-defined rules, we construct a dataset of JD-Talent pairs, predominantly consisting of positive samples. The resulting dataset serves as the foundation for model training.

To ensure high-quality researcher profiles, authors are filtered based on:
Having published at least two papers in the past ten years (2014-2023), and the citation count of each paper should exceed the number of years since its publication.
For each author, we randomly select up to 50 articles published within the past five years. This ensures that the dataset reflects the author’s most recent academic activity.
From each selected article, we extract the associated keywords, title, and abstract.
Each author's most recent institution, as identified in their latest publication, was used to determine their current location.
This information is useful for city-based matching with JDs.
The academic seniority of an author can be calculated starting from the year of their first published article.

Authors are matched with JD based primarily on the location, academic seniority, and research focus of the author and JD.
We collected 17,830 JD samples and matched them with 19,246 researchers. Each JD contains job-related information, such as required skills and responsibilities, with an average length of 554 words. For the matched authors, we combined the keywords, title, and abstract from their published articles into profiles, with an average length of 4,139 words.
The dataset includes a total of 79,513 JD-Talent pairs, meaning that, on average, each JD is matched to approximately 4.5 authors.
Table~\ref{tab:jd_talent_summary} summarizes the key statistics of the dataset.

\begin{table}[tb]
\centering
\renewcommand{\arraystretch}{1.5} 
\begin{tabular}{@{}lccc@{}}
\toprule
\textbf{Data} & \textbf{Count} & \textbf{Mean Word Count} \\ 
\midrule
JD                & 17,830                & 554             \\ 
Author            & 19,246                & 4,139           \\ 
\bottomrule
\end{tabular}
\caption{Key statistics of the JD-Talent dataset.}
\label{tab:jd_talent_summary}
\end{table}

\section{Method}

\subsection{Contrastive Learning}

Contrastive learning is a self-supervised learning paradigm that aims to learn representations by pulling similar samples closer while pushing dissimilar samples apart in the embedding space~\citep{chen2020simple, khosla2020supervised, robinson2021contrastive, gao2021simcse, liu2021simcls}.
Given a query sample \( \mathbf{x}_i \), a positive sample \( \mathbf{x}_i^+ \) (e.g. an augmented version of \( \mathbf{x}_i \) or a labeled positive sample), and negative samples \( \mathbf{x}_j^- \), the goal is to maximize the similarity between \( \mathbf{x}_i \) and \( \mathbf{x}_i^+ \), while minimizing the similarity between \( \mathbf{x}_i \) and \( \mathbf{x}_j^- \).
This approach is particularly effective in tasks where explicit supervision can be utilized to improve the embedding of the learned representations.

The InfoNCE loss, commonly used in contrastive learning, is defined as:
\begin{equation}
\mathcal{L}_{i} = -\log \frac{\exp(\text{sim}(\mathbf{z}_i, \mathbf{z}_i^+) / \tau)}{\sum_{j=1}^N \exp(\text{sim}(\mathbf{z}_i, \mathbf{z}_j) / \tau)},
\end{equation}
where 
$\mathbf{z}_i$ and $\mathbf{z}_i^+$ are the embeddings of the query and positive samples, respectively,
$\tau$ is the temperature hyperparameter,
$\text{sim}(\mathbf{z}_i, \mathbf{z}_j)$ denotes the similarity function, often defined as cosine similarity.

By explicitly using label information, contrastive learning encourages matched samples to form tighter clusters in the embedding space while maintaining separation between unmatched samples.
In practice, we followed the procedures and recommendations in M3-Embedding~\citep{chen2024bge} to implement the contrastive learning approach.

The contrastive learning step can be iterated multiple times to refine the model's ability to distinguish positive and negative samples in a curriculum learning manner~\cite{bengio2009curriculum}. 
A particularly effective strategy during this process is the incorporation of hard negatives—examples that are challenging for the model to differentiate from positives because of their similarity.
Hard negatives push the model to focus on subtle, fine-grained distinctions, which enhances its ability to learn more meaningful representations.

\subsection{Preference Optimization}

\paragraph{Reinforcement Learning from Human Feedback.}

Reinforcement Learning from Human Feedback (RLHF) is a framework for aligning machine learning models with human preferences~\citep{ziegler2019finetuning, stiennon2020learning, bai2022training, bai2022constitutional, ouyang2022training, zheng2023secrets}. 
RLHF pipeline generally involves three main steps: 
(1) supervised fine-tuning, (2) training a reward model, and (3) reinforcement learning-based fine-tuning.
During the RL fine-tuning phase,
the optimization objective is to maximize the expected reward under the model's policy~\citep{ziegler2019finetuning, ouyang2022training, stiennon2020learning, hsu2020revisiting}:

\begin{equation}\label{eq:RL}
\max_{\pi_{\theta}} \  \mathbb{E}_{x\sim \mathcal{D}, y\sim \pi_{\theta}(y \mid x)} \\
\ \bigl[r_{\phi}(x, y) - 
\beta {\frac{\pi_{\theta}(y\mid x)}{\pi_\text{ref}(y\mid x)}}\bigr],
\end{equation}

\noindent where $\mathcal{D}$ is the prompt dataset,
$\pi_\theta$ is the model's policy parameterized by $\theta$,
$\pi_\text{ref}$ is the reference policy (typically set to be the supervised fine-tuned model before the RL phase),
$r_{\phi}(x, y)$ denotes the reward given by the reward model,
and
$\beta$ is the KL penalty coefficient that prevents $\pi_\theta$ from moving too far from the reference policy.

While RLHF has proven effective in aligning language models with human preferences, it has several limitations~\citep{rafailov2023direct, skalse2022defining, zheng2023secrets}. One key challenge is the complexity and instability of reinforcement learning algorithms, such as Proximal Policy Optimization (PPO)~\citep{schulman2017proximal}, which are commonly used to optimize the model. 
These methods are also computationally expensive and sensitive to hyperparameters, requiring careful tuning to achieve stable results~\citep{huang2024the}. 
Additionally, RLHF relies heavily on the quality of the reward model, 
and over-optimization toward the reward function may lead to unintended behaviors, such as mode collapse or exploitation of weaknesses in the reward model~\citep{skalse2022defining, wang2024secrets}.

\paragraph{Direct Preference Optimization (DPO).}  

Unlike traditional RLHF approaches, DPO~\citep{rafailov2023direct} directly optimizes a model’s parameters based on preference data
and eliminates the need for a separate reward model and reinforcement learning phase,
making it computationally efficient and easier to implement.

DPO relies on human or AI feedback in the form of pairwise comparisons, where one output is preferred over another.
By reparameterizing the reward function using the optimal policy,
\citet{rafailov2023direct}
derived the pairwise DPO preference loss:

\begin{equation}
\label{eq:dpo_loss}
\begin{aligned}
&\mathcal{L}_\text{DPO}(\pi_{\theta}) = -\mathbb{E}_{(x, y_w, y_l)\sim \mathcal{D}} \\
&\log \sigma \left(\beta \log {\frac{\pi_{\theta}(y_w\mid x)}{\pi_\text{ref}(y_w\mid x)}} - \beta \log {\frac{\pi_{\theta}(y_l\mid x)}{\pi_\text{ref}(y_l\mid x)}}\right),
\end{aligned}
\end{equation}

\noindent where
$\mathcal{D}$ is the preference dataset,
$x$ is the prompt, 
$y_w$ is the winning response (preferred),
$y_l$ is the losing response (not preferred),
and 
$\sigma$ is the sigmoid function.

\subsection{Rank Preference Optimization (RankPO)}

RankPO is a tailored extension of DPO, specifically designed to address the unique challenges of ranking and retrieval tasks, such as job-talent matching, where a query (e.g., a job description) needs to be compared against a large pool of candidates (e.g., profiles). 
RankPO adapts the preference optimization process to operate efficiently in embedding spaces for large-scale ranking problems.

Given a pairwise comparison between candidates 
$y_w \succ y_l$
for a query $x$, 
RankPO optimizes the model to ensure that the embedding similarity
$\text{sim}(\mathbf{z}_x,\mathbf{z}_y^w)$ is greater than $\text{sim}(\mathbf{z}_x,\mathbf{z}_y^l)$, 
where 
$\mathbf{z}_y^{w}$ is the embedding of $y_{w}$ 
and
$\text{sim}(\mathbf{z}_x,\mathbf{z}_y)$ denotes the similarity score between $x$ and $y$.
Note that,
given a query $x$, and two candidates $y_w$ and $y_l$, 
the probability of $y_w$ being selected under policy $\pi$ is:

\begin{equation}
P^{\pi}(y_w \mid x) = \frac{\exp(\text{sim}^{\pi}(\mathbf{z}_x,\mathbf{z}_y^w)/\tau)}{\sum_{y \in \{y_w, y_l\}}\exp(\text{sim}^{\pi}(\mathbf{z}_x,\mathbf{z}_y)/\tau)},
\end{equation}

\noindent where $\tau$ is the temperature parameter.
The logratio of winning and losing responses or candidates under the same policy $\pi$ would be:

\begin{equation}
\label{eq:rankpo_logratio}
\begin{aligned}
& \log {\frac{P^{\pi}(y_w \mid x)}{P^{\pi}(y_l \mid x)}} = \log {\frac{\exp(\text{sim}^{\pi}(\mathbf{z}_x,\mathbf{z}_y^w)/\tau)}{\exp(\text{sim}^{\pi}(\mathbf{z}_x,\mathbf{z}_y^l)/\tau)}} \\
& \quad = \text{sim}^{\pi}(\mathbf{z}_x,\mathbf{z}_y^w)/\tau - \text{sim}^{\pi}(\mathbf{z}_x,\mathbf{z}_y^l)/\tau . 
\end{aligned}
\end{equation}

Building upon the DPO loss introduced in Eq.~\ref{eq:dpo_loss}, 
we generalize the loss function using a term $f$ to represent various possible forms of the loss function (e.g., negative log-likelihood, hinge loss)\footnote{Note that, after generalizing the loss function to a broader form represented as $f$,
the first negative symbol in Eq.~\ref{eq:dpo_loss} is now absorbed into the definition of $f$.}. Using this generalization, we can derive the RankPO loss as follows:

\begin{equation}
\label{eq:rankpo_loss}
\begin{aligned}
& \mathcal{L}_\text{RankPO}(\pi_{\theta}) 
= \mathbb{E}_{(x, y_w, y_l)\sim \mathcal{D}} \\
& \quad f \left(
\beta \log {\frac{P^{\pi_{\theta}}(y_w \mid x)}{P^{\pi_{\theta}}(y_l\mid x)}} - 
\beta \log {\frac{P^{\pi_{\text{ref}}}(y_w\mid x)}{P^{\pi_{\text{ref}}}(y_l\mid x)}}
\right) \\
& = \mathbb{E}_{(x, y_w, y_l)\sim \mathcal{D}} \ \big[\ f \big( \\&
\ \ \ \ \ \  \beta \cdot \text{sim}^{\pi_{\theta}}(\mathbf{z}_x,\mathbf{z}_y^w)/\tau 
- \beta \cdot \text{sim}^{\pi_{\theta}}(\mathbf{z}_x,\mathbf{z}_y^l)/\tau \\&
\ - \beta \cdot \text{sim}^{\pi_{\text{ref}}}(\mathbf{z}_x,\mathbf{z}_y^w)/\tau 
+ \beta \cdot \text{sim}^{\pi_{\text{ref}}}(\mathbf{z}_x,\mathbf{z}_y^l)/\tau \ \big) \big].
\end{aligned}
\end{equation}

\noindent 
This will directly align the embeddings with the pairwise preference dataset, pushing preferred candidates ranked higher during training.

Inspired by the SimPO framework~\citep{meng2024simpo}, we can omit the reference model or policy, resulting in a simplified version of RankPO, which we term SimRankPO.
The loss function for SimRankPO can be written as:

\begin{equation}
\label{eq:simrankpo_loss}
\begin{aligned}
& \mathcal{L}_\text{SimRankPO}(\pi_{\theta}) 
= \mathbb{E}_{(x, y_w, y_l)\sim \mathcal{D}} \ \big[\ f \big( \\&
\ \ \ \ \ \  \beta \cdot \text{sim}^{\pi_{\theta}}(\mathbf{z}_x,\mathbf{z}_y^w)/\tau 
- \beta \cdot \text{sim}^{\pi_{\theta}}(\mathbf{z}_x,\mathbf{z}_y^l)/\tau 
\ \big) \big].
\end{aligned}
\end{equation}


\begin{table}[b]
\centering
\begin{tabular}{ll}
\toprule
\textbf{Hyperparameter}        & \textbf{Value}          \\
\midrule
Number of epochs               & 3                       \\
Batch size                     & 8                       \\
Number of negatives            & 5                       \\
Base model                     & llama-3.2-1b            \\
Optimizer                      & AdamW                   \\
Learning rate                  & 1e-5                    \\
Learning rate scheduler        & cosine                  \\
Warmup ratio                   & 0.1                    \\
Temperature                    & 0.02                    \\
Max token length (JD)          & 1280               \\
Max token length (Talent)      & 4096              \\
\bottomrule
\end{tabular}
\caption{Training setup and hyperparameters in contrastive learning stage.}
\label{tab:cl_setup}
\end{table}


\begin{table*}[htb]
\centering
\begin{tabular}{llcccccc}
\toprule
\multirow{2}{*}{\textbf{Strategy}} & \multirow{2}{*}{\textbf{Training Data}}
& \multicolumn{2}{c}{\textbf{MRR@k}} 
& \multicolumn{2}{c}{\textbf{Recall@k}} 
& \multicolumn{2}{c}{\textbf{nDCG@k}} \\
\cmidrule(r){3-4} \cmidrule(r){5-6} \cmidrule(r){7-8}
& & \textbf{5} & \textbf{20} 
& \textbf{5} & \textbf{20} 
& \textbf{5} & \textbf{20} \\
\midrule
\multirow{3}{*}{Curriculum}
& {random} & 0.605 & 0.617 & 0.560 & 0.704 & 0.533 & 0.616 \\
 & {hn1} & 0.676 & 0.686 & 0.622 & 0.727 & 0.606 & 0.677 \\
 & {hn2} & 0.696 & 0.705 & 0.622 & 0.719 & 0.617 & 0.684 \\
\midrule
\multirow{2}{*}{Combined} & {random, hn1} & 0.692 & 0.702 & 0.643 & 0.763 & 0.623 & 0.694 \\
& {random, hn1, hn2} & \textbf{0.705} & \textbf{0.712} & \textbf{0.657} & \textbf{0.766} & \textbf{0.638} & \textbf{0.706} \\
\bottomrule
\end{tabular}
\caption{Performance of contrastive learning across iterations. 
The table presents model performance on three metrics: MRR, Recall, and nDCG, at different cutoffs $k$ (5 and 20) from a single run. 
The best result in each column is highlighted in bold.
The notation \texttt{random} refers to training with random negatives, whereas \texttt{hn1} and \texttt{hn2} indicate training with hard negative set 1 and set 2, respectively. 
The combined datasets (e.g., \texttt{random, hn1}) use data accumulated from multiple iterations.
"Curriculum" refers to curriculum learning which sequentially trains the model over iterations, while "Combined" refers to training the base model on the combined dataset.
}
\label{tab:contrastive_results}
\end{table*}

\section{Experiments}

The experiments mainly contain two stages:
contrastive learning and rank preference optimization.
We didn't make any special model selection but directly used the last checkpoint from each training phase as the trained model for use.
Below, we detail the experimental setup
and results for each stage.

\subsection{Contrastive Learning}

\subsubsection{Data and Setup}

For contrastive learning, the dataset consists of JDs as queries and talent profiles as candidates. 
%
Among the  constructed dataset, 90\% (16,047 JDs) were allocated to the training dataset, while 10\% (1,783 JDs) were allocated to the testing dataset.

We used llama-3.2-1b~\citep{dubey2024llama} as the base model, a lightweight version of the llama family with about 1.2 billion parameters.
This model provided the foundational embeddings for JDs and talent profiles.
The training was conducted using 4 NVIDIA A100 80GB GPUs. 
Key hyperparameters used in the experiments can be found in Table~\ref{tab:cl_setup}.

For each query (a JD), 
we randomly selected one positive sample (a matched talent profile) and five negative samples, which were either random or hard negatives. During training, in-batch negatives and cross-device negatives were used to enhance the contrastive learning process. The FAISS library~\citep{douze2024faiss} was used for efficient embedding search.

The training process followed a curriculum learning strategy over three iterations. 
In the first iteration, only random negative samples were used to establish a baseline. 
In subsequent iterations, hard negatives, mined from the previous iteration, were used to progressively improve the model's ability to distinguish challenging cases.

\subsubsection{Results}

The performance of the trained models was evaluated on the test dataset using three standard ranking metrics: Mean Reciprocal Rank (MRR), Recall@k, and Normalized Discounted Cumulative Gain (nDCG@k). These metrics provide a comprehensive assessment of the model's ranking and retrieval quality.

\paragraph{Curriculum learning.}
Table~\ref{tab:contrastive_results} summarizes the model performance on the test dataset across three training iterations, with results reported for MRR@k, Recall@k, and nDCG@k (See Table~\ref{tab:full_contrastive_results} for further details). The results demonstrate that incorporating hard negatives through curriculum learning leads to steady improvements in ranking performance across iterations. 
For example,
nDCG@20 increased significantly from 0.616 in iteration 1 to 0.684 in iteration 3, reflecting improved ranking quality as the model was exposed to harder negatives.
The improvement from iteration 2 to iteration 3, however, was marginal compared to that from iteration 1 to iteration 2, indicating diminishing returns as the model may approach its performance ceiling under the current training regime.

\paragraph{Combined training vs. curriculum learning.}

While curriculum learning provided consistent improvements across iterations, the best results were achieved by fine-tuning the base model on the combined dataset, which includes all negatives from the three iterations (random negatives, hard negative sets 1 and 2). 
Notably,
the model demonstrated significant improvements in Recall, particularly at higher $k$. For instance, Recall@20 improved from 0.719 in iteration 3 to 0.766 under combined training. 
Moreover, training the base model on a combined dataset consisting of random negatives and hard negative set 1 achieved superior performance on both Recall and nDCG compared to iteration 3 of curriculum learning.
These results demonstrate that combining negatives from all iterations enables the model to better generalize and retrieve relevant candidates, overcoming the potential limitations of incremental training in curriculum learning.

\begin{figure*}[t]
  \centering
  \includegraphics[width=\textwidth]{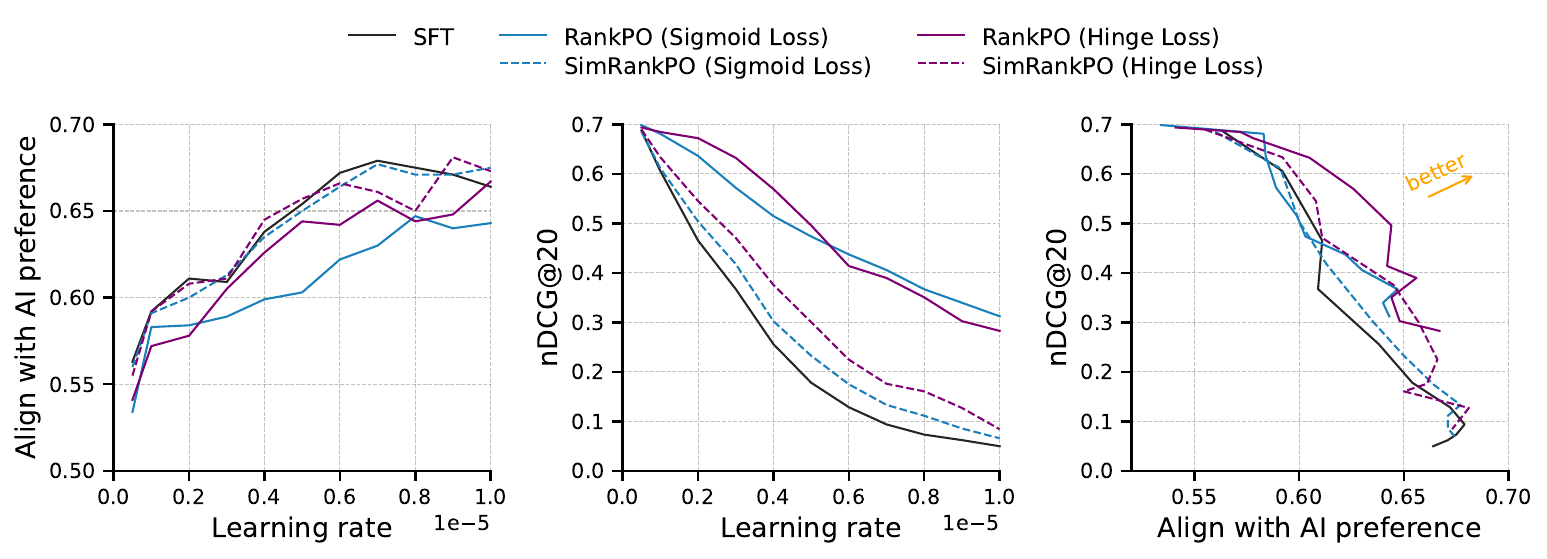}
  \caption {Comparison of RankPO and SFT across different learning rates in terms of Agreement with AI and nDCG@20. 
  Lines represent the average results over two random seeds.
  Higher values 
  indicate better adaptation to AI preferences and better retention of previously learned capabilities.
  It highlights RankPO's superior ability to 
  maintain previously learned capabilities at higher alignment levels compared to SFT.
  }
  \label{fig:rankpo_performance}
\end{figure*}

\subsection{Rank Preference Optimization}


\subsubsection{Preference Data Annotation}

Given the high cost and difficulty of obtaining expert annotations  for ranking preferences, 
we leveraged advanced AI systems
to annotate pairwise preferences between candidates.
Specifically, DeepSeek v3~\citep{liu2024deepseek}
was used to generate pairwise rankings based on textual content.
The prompt used can be found in Appendix Figure~\ref{fig:prompt}.

More than 3,000 JDs were randomly selected from the training dataset. For each JD, the model from the contrastive learning phase (trained in the combined datasets) was first applied to rank all candidates. 
From the top 100 ranked candidates, two were randomly selected (with the order randomly shuffled) for pairwise comparison, and the AI system estimated which candidate should be ranked higher. 
For evaluation, a separate set of 500 JDs was randomly selected from the test dataset,
and pairwise preferences were annotated using the same process, except that the pairs are randomly selected from the top 20 ranked candidates.
This annotated test set was designed to evaluate the model’s alignment with AI-inferred preferences.
The alignment with AI preferences is simply measured as the proportion of correct predictions made by the RankPO model compared to the labels annotated by the AI system.

\subsubsection{Experiment Setup}

For RankPO and SimRankPO, we implement two types of loss functions: negative log-sigmoid (denoted as \textit{sigmoid loss}) and hinge loss. 
For comparison,
Supervised Fine-Tuning (SFT)
is used as the baseline in this study.
Specifically,
SFT is implemented by directly optimizing the cross-entropy loss on the temperature-scaled similarity score between input pairs, denoted as $\text{sim}(\mathbf{z}_x,\mathbf{z}_y)/\tau$.

Unlike RankPO, SFT and SimRankPO don't include a reference model constraint to regularize updates or mitigate forgetting. 
This lack of constraints may allow them to quickly adapt to new data, 
but this rapid adaptation often comes at the cost of forgetting previously learned capabilities.
In addition,
the comparison between SimRankPO and RankPO in this study resembles the contrast between SimPO and DPO in prior research~\citep{meng2024simpo}, except that the SimRankPO model here doesn't have a constraint like response length in SimPO.

Training in this stage was conducted using 1 NVIDIA A100 80GB GPU. 
Temperature $\tau$ is set to 0.1, and the KL penalty coefficient $\beta$ is set to 2.0.
Other key hyperparameters used in the experiments are the same with the previous stage (see Table~\ref{tab:rankpo_setup} in Appendix).

\subsubsection{Results}

To evaluate the performance of RankPO, SimRankPO and SFT, we conducted experiments under varying learning rates.
%
Figure~\ref{fig:rankpo_performance} illustrates 
the alignment with AI preference
and
nDCG@20 as a function of the learning rate,
as well as the direct comparison between 
nDCG@20 and the alignment with AI preference.
Detailed results can be found in Appendix Table~\ref{tab:rankpo_performance} and Table~\ref{tab:simrankpo_performance}.

The results indicate that SFT is more sensitive to the choice of learning rate than RankPO. 
When the learning rate is small, both RankPO and SFT demonstrate better retention of previously learned abilities. This is reflected in higher nDCG@20 scores at lower learning rates, suggesting that smaller learning rates help to reduce catastrophic forgetting during fine-tuning.

\paragraph{SFT achieves higher alignment but suffers from severe forgetting.}
SFT is able to achieve higher alignment with AI preference, 
which 
is consistent
with expectations
given that it directly optimizes for the new objective without constraints. However, this rapid adaptation leads to significant forgetting of prior capabilities, as evidenced by lower nDCG@20 scores. 
Similar phenomenon can also be observed in the comparisons between SimRankPO and their RankPO counterparts.
This behavior highlights the inherent conflict between fine-tuning for alignment and retaining generalization capabilities learned during contrastive learning.

\paragraph{RankPO balances alignment and prior knowledge retention.}
While RankPO also experiences some degree of forgetting, its rate of degradation is much slower than that of SFT. This indicates that RankPO provides a better balance between aligning with AI preferences and preserving previously learned abilities (Figure~\ref{fig:rankpo_performance}, right panel). 
Such a balance is particularly valuable
in practical applications that require both generalization and alignment with new data.

A particularly notable finding is that RankPO significantly outperforms SFT at higher levels of alignment with AI preference. 
For example,
at an alignment score of 0.647, RankPO (with sigmoid loss) achieves nDCG@20 = 0.367, 
whereas SFT only gets nDCG@20 = 0.256 at a similar alignment level of 0.638.
Furthermore, with an alignment score of 0.656, RankPO (with hinge loss) even achieves nDCG@20 = 0.390. 
Overall, we can also find that hinge loss generally performs better than sigmoid loss for RankPO in the experiments.

\paragraph{Practical implications.}
The results reveal a critical trade-off between alignment with AI or human preference and the retention of previously learned knowledge~\citep{bai2022training}.
These findings highlight that the choice between RankPO and SFT should depend on the specific requirements of the task:
SFT is well-suited for tasks when adaptation to new data is the primary goal.
RankPO, on the other hand, 
provides a more balanced approach, making it ideal for tasks that require both strong alignment with human or AI preferences and the preservation of previously learned knowledge.

In this context, SimRankPO can be viewed as a middle ground between SFT and RankPO—a compromise that balances the strengths of both methods. 
It provides better retention than SFT while can achieve stronger alignment than RankPO in certain cases.

\section{Related Work}

\subsection{Contrastive Learning}
Contrastive learning has emerged as a powerful technique for tasks requiring fine-grained alignment between two entities, such as image-text matching and question-answer retrieval~\citep{chen2020simple, khosla2020supervised, robinson2021contrastive, gao2021simcse, liu2021simcls}. Supervised contrastive learning, in particular, has shown promise in scenarios where labeled positive and negative pairs are available. Our work adapts this paradigm to the JD-Talent matching task, with iterative improvements using hard negatives in a curriculum learning manner~\cite{bengio2009curriculum} and a combined training approach.

\subsection{Preference Optimization}
Preference-based learning methods, such as (RLHF~\citep{ziegler2019finetuning, stiennon2020learning, bai2022training, bai2022constitutional, ouyang2022training, zheng2023secrets} and DPO~\citep{rafailov2023direct}, have been successfully applied in domains like dialogue generation and summarization. However, their application to recommendation tasks remains under-explored~\citep{chen2024on}. 
Inspired by DPO, we introduce RankPO, a preference-based optimization method tailored for ranking tasks like JD-Talent matching and beyond.

\section{Conclusion}

In this work, we proposed and evaluated a two-stage framework for fine-tuning models to align with AI preferences while retaining previously learned knowledge. 
The first stage leverages contrastive learning to train the model on 
rule-based data.
The second stage focuses on optimizing the model to align with specific AI preferences using the proposed RankPO method. 

Our results show that the proposed framework effectively balances alignment with AI preferences and retention of previously learned capabilities. Contrastive learning plays a critical role by providing a strong backbone for ranking tasks, while the preference optimization stage refines the model to meet specific alignment goals. 
These findings highlight the flexibility of the system in adapting to various alignment objectives while preserving generalization.

\section*{Limitations}

Several limitations should be noted:

\noindent
\textbf{AI preference as a non-gold standard.}
The alignment process relies on AI-generated preferences as a proxy for desired objectives. However, these outputs may not always represent a true gold standard and can sometimes embed biases or inaccuracies. Nevertheless, our framework is adaptable to other sources of supervision, such as human feedback or task-specific metrics, which could further improve alignment quality.

\noindent
\textbf{Task-specific evaluation.}
Our experiments and evaluations were conducted exclusively on the job-talent matching task.
While our framework is conceptually applicable to other similar tasks, such as recommendation systems, ranking problems, or matching-based applications, further studies are needed to validate its effectiveness in broader contexts.

\noindent
\textbf{Dependency on pre-trained models.}
The performance of the framework depends heavily on the quality of the pre-trained model and the contrastive learning stage. Suboptimal pre-training or contrastive learning may impact downstream performance. Further research is needed to explore the robustness of the framework under different pre-training and post-training strategies.

\noindent
\textbf{Computational cost.}
Compared with SFT,
RankPO introduces additional computational overhead, which may pose challenges for scaling in resource-limited environments. 
Optimizing the computational efficiency of the framework is an important direction for future work.


\section*{Acknowledgments}

The work described in this paper was partially supported by InnoHK initiative, The Government of the HKSAR, and Laboratory for AI-Powered Financial Technologies.

\bibliography{custom}

\clearpage 


\appendix




\section{Training Setup}

Training setup in the RankPO stage.

\begin{table}[htb]
\centering
\begin{tabular}{ll}
\toprule
\textbf{Hyperparameter}        & \textbf{Value}         \\
\midrule
Number of epochs               & 3                      \\
Batch size                     & 8                      \\
Optimizer                      & AdamW                   \\
Learning rate scheduler        & cosine                 \\
Warmup ratio                   & 0.1                    \\
Temperature                    & 0.1                    \\
Beta                           & 2.0                    \\
Max token length (JD)          & 1280                   \\
Max token length (Talent)      & 4096                   \\

\bottomrule
\end{tabular}
\caption{Training setup and hyperparameters in RankPO optimization.}
\label{tab:rankpo_setup}
\end{table}

\section{Metrics for Evaluation}

\label{sec:metrics}

\subsection{Mean Reciprocal Rank (MRR)}

Mean Reciprocal Rank (MRR) is a commonly used metric for evaluating ranking systems. 
It's the average of the reciprocal ranks of the first relevant item across all queries.

\[
\text{MRR} = \frac{1}{N} \sum_{i=1}^N \frac{1}{\text{rank}_i}
\]

\noindent
where   
\(N\) is the total number of queries.
\(\text{rank}_i\) is the rank position of the first relevant item for the \(i\)-th query.

A higher MRR indicates that relevant items generally appear earlier in the ranked results.

\subsection{Recall}

Recall measures the fraction of relevant items that are successfully retrieved by the model. It is particularly useful when evaluating how well a system captures all relevant results, regardless of their position in the ranking.

\[
\text{Recall@k} = \frac{\text{\#Relevant items retrieved in top } k}{\text{\#All relevant items}}
\]

\noindent
where \(k\) is the cutoff rank.

Recall ranges from 0 to 1, with 1 indicating that all relevant items are retrieved within the top \(k\) positions.

\subsection{Normalized Discounted Cumulative Gain (nDCG)}

Normalized Discounted Cumulative Gain (nDCG) is a ranking metric that evaluates the quality of the ranked results by considering the position of relevant items. It assigns higher importance to relevant items appearing earlier in the ranking. The "normalized" part ensures the score is scaled between 0 and 1.

First, the Discounted Cumulative Gain (DCG) is computed:

\[
\text{DCG@k} = \sum_{i=1}^k \frac{\text{rel}_i}{\log_2(i + 1)}
\]

\noindent
where
\(k\) is the cutoff rank,
\(\text{rel}_i\) is the relevance score of the item at rank \(i\).

Next, the DCG is normalized by the Ideal DCG (IDCG), which is the maximum possible DCG achievable by a perfect ranking:

\[
\text{nDCG@k} = \frac{\text{DCG@k}}{\text{IDCG@k}}
\]

nDCG ranges from 0 to 1, where 1 represents a perfectly ranked list.

\section{Contrastive Learning Results}

Table~\ref{tab:full_contrastive_results} presents the results of the contrastive learning stage
on three metrics: MRR, Recall, and nDCG, at different cutoffs $k$ (5, 10, 20, and 100).
This is an extension of Table~\ref{tab:contrastive_results} in the main text.

\begin{table*}[htb]
\centering
\begin{tabularx}{\textwidth}{>{\raggedright\arraybackslash}l 
>{\centering\arraybackslash}X >{\centering\arraybackslash}X >{\centering\arraybackslash}X >{\centering\arraybackslash}X 
>{\centering\arraybackslash}X >{\centering\arraybackslash}X >{\centering\arraybackslash}X >{\centering\arraybackslash}X 
>{\centering\arraybackslash}X >{\centering\arraybackslash}X >{\centering\arraybackslash}X >{\centering\arraybackslash}X}
\toprule
\multirow{2}{*}{\textbf{Train Data}}
& \multicolumn{4}{c}{\textbf{MRR@k}} 
& \multicolumn{4}{c}{\textbf{Recall@k}} 
& \multicolumn{4}{c}{\textbf{nDCG@k}} \\
\cmidrule(r){2-5} \cmidrule(r){6-9} \cmidrule(r){10-13}
& \textbf{5} & \textbf{10} & \textbf{20} & \textbf{100} 
& \textbf{5} & \textbf{10} & \textbf{20} & \textbf{100} 
& \textbf{5} & \textbf{10} & \textbf{20} & \textbf{100} \\
\midrule
\texttt{random} & 0.605 & 0.614 & 0.617 & 0.619 & 0.560 & 0.625 & 0.704 & 0.811 & 0.533 & 0.575 & 0.616 & 0.664 \\
\texttt{hn1} & 0.676 & 0.684 & 0.686 & 0.687 & 0.622 & 0.669 & 0.727 & 0.814 & 0.606 & 0.645 & 0.677 & 0.718 \\
\texttt{hn2} & 0.696 & 0.704 & 0.705 & 0.707 & 0.622 & 0.663 & 0.719 & 0.808 & 0.617 & 0.652 & 0.684 & 0.725 \\
\midrule
\texttt{random+hn1} & 0.692 & 0.699 & 0.702 & 0.703 & 0.643 & 0.689 & 0.763 & 0.856 & 0.623 & 0.656 & 0.694 & 0.735 \\
\texttt{random+hn1+hn2} & 0.705 & 0.710 & 0.712 & 0.713 & 0.657 & 0.699 & 0.766 & 0.853 & 0.638 & 0.669 & 0.706 & 0.744 \\
\bottomrule
\end{tabularx}
\caption{Performance of contrastive learning across iterations. 
The table reports model performance on three metrics: MRR, Recall, and nDCG, at different cutoffs $k$ (5, 10, 20, and 100). 
The notation \texttt{random} refers to training with random negatives, whereas \texttt{hn1} and \texttt{hn2} indicate training with hard negative sets 1 and 2, respectively. 
The combined datasets (e.g., \texttt{random, hn1}) use data accumulated from multiple iterations.
"Curriculum" refers to curriculum learning, which sequentially trains the model over iterations, while "Combined" refers to training the base model on the combined dataset.
}
\label{tab:full_contrastive_results}
\end{table*}

\begin{figure*}[htb]
\begin{tcolorbox}[
    colback=gray!5!white,    
    colframe=gray!70!black,  
    boxrule=0.5mm,           
    rounded corners,         
    fonttitle=\bfseries,     
    ]

You are a professional recruitment expert in academic areas.\\

I will provide you with a job description for a specific role and resumes/lists of articles from two candidates (keywords, title and abstract from the articles). \\

Your task is to:\\
1. Analyze each candidate's compatibility with the job description.\\
2. Highlight the strengths and weaknesses of each candidate.\\
3. Assign a compatibility score (out of 100) to each candidate.\\
4. Recommend the most suitable candidate and explain the reasoning behind your recommendation.\\

<Job Description>\\
\{JD\}\\

<Candidate A>\\
\{Candidate A's profile\}\\

<Candidate B>\\
\{Candidate B's profile\}\\

Your response should use the format:\\
Comparison: <one-sentence comparison and explanation>\\
Preferred: <"A" or "B">\\
Confidence score: <100>
\end{tcolorbox}
\caption{Prompt used for AI annotation.}
\label{fig:prompt}
\end{figure*}

\section{Prompt Used for AI Annotation}

\label{sec:prompt}

The prompt used for AI-based annotation during the experimental process is shown in Figure~\ref{fig:prompt}. 
The prompt was designed to guide the AI model in generating preference annotations for the provided candidate pairs based on the content. 

\section{Preference Optimization Results}

\label{sec:preference_optimization}

Table~\ref{tab:rankpo_performance} and Table~\ref{tab:simrankpo_performance} presents the evaluation results after applying Rank Preference Optimization (RankPO). 
This evaluation assesses the model's ability to retain capabilities acquired through contrastive learning (MRR, Recall, nDCG), while also measuring its alignment with the AI system. 
The alignment with AI preferences is measured as the proportion of correct predictions made by the RankPO model compared to the labels annotated by the AI system.

\begin{table*}[htb]
\centering
\begin{tabular}{llcccccccccc}
\toprule
\multirow{2}{*}{\textbf{Method}} & \multirow{2}{*}{\textbf{LR}} & \multirow{2}{*}{\textbf{Align.}} 
& \multicolumn{3}{c}{\textbf{MRR@k}} 
& \multicolumn{3}{c}{\textbf{Recall@k}} 
& \multicolumn{3}{c}{\textbf{nDCG@k}} \\
\cmidrule(r){4-6} \cmidrule(r){7-9} \cmidrule(r){10-12}
&  &   & \textbf{5} & \textbf{20} & \textbf{100} 
& \textbf{5} & \textbf{20} & \textbf{100} 
& \textbf{5} & \textbf{20} & \textbf{100} \\
\midrule
\multirow{11}{*}{SFT} 
& 5e-7 & 0.563 & 0.692 & 0.701 & 0.702 & 0.639 & 0.739 & 0.833 & 0.623 & 0.687 & 0.729 \\
& 1e-6 & 0.592 & 0.623 & 0.632 & 0.634 & 0.548 & 0.648 & 0.761 & 0.542 & 0.606 & 0.654 \\
& 2e-6 & 0.611 & 0.483 & 0.496 & 0.499 & 0.393 & 0.501 & 0.650 & 0.399 & 0.465 & 0.526 \\
& 3e-6 & 0.609 & 0.377 & 0.392 & 0.395 & 0.289 & 0.390 & 0.545 & 0.304 & 0.367 & 0.431 \\
& 4e-6 & 0.638 & 0.248 & 0.262 & 0.267 & 0.186 & 0.268 & 0.424 & 0.203 & 0.256 & 0.320 \\
& 5e-6 & 0.654 & 0.159 & 0.174 & 0.178 & 0.118 & 0.185 & 0.322 & 0.133 & 0.178 & 0.237 \\
& 6e-6 & 0.672 & 0.107 & 0.119 & 0.123 & 0.074 & 0.134 & 0.249 & 0.090 & 0.129 & 0.182 \\
& 7e-6 & 0.679 & 0.075 & 0.086 & 0.090 & 0.050 & 0.094 & 0.201 & 0.062 & 0.094 & 0.142 \\
& 8e-6 & 0.675 & 0.056 & 0.065 & 0.069 & 0.037 & 0.071 & 0.162 & 0.047 & 0.073 & 0.116 \\
& 9e-6 & 0.671 & 0.047 & 0.054 & 0.057 & 0.031 & 0.059 & 0.136 & 0.040 & 0.062 & 0.099 \\
& 1e-5 & 0.664 & 0.035 & 0.041 & 0.044 & 0.024 & 0.046 & 0.112 & 0.031 & 0.050 & 0.083 \\
\midrule
\multirow{11}{*}{\makecell{RankPO \\ (Sigmoid)}}
& 5e-7 & 0.534 & 0.700 & 0.709 & 0.710 & 0.650 & 0.757 & 0.846 & 0.632 & 0.699 & 0.738 \\
& 1e-6 & 0.583 & 0.687 & 0.696 & 0.698 & 0.629 & 0.739 & 0.835 & 0.615 & 0.681 & 0.723 \\
& 2e-6 & 0.584 & 0.648 & 0.658 & 0.660 & 0.586 & 0.695 & 0.809 & 0.569 & 0.636 & 0.685 \\
& 3e-6 & 0.589 & 0.587 & 0.599 & 0.601 & 0.510 & 0.639 & 0.772 & 0.497 & 0.572 & 0.627 \\
& 4e-6 & 0.599 & 0.525 & 0.539 & 0.542 & 0.449 & 0.589 & 0.735 & 0.435 & 0.515 & 0.574 \\
& 5e-6 & 0.603 & 0.480 & 0.495 & 0.498 & 0.407 & 0.547 & 0.708 & 0.394 & 0.473 & 0.538 \\
& 6e-6 & 0.622 & 0.441 & 0.457 & 0.460 & 0.367 & 0.507 & 0.677 & 0.359 & 0.437 & 0.503 \\
& 7e-6 & 0.630 & 0.402 & 0.418 & 0.421 & 0.337 & 0.476 & 0.644 & 0.328 & 0.406 & 0.472 \\
& 8e-6 & 0.647 & 0.358 & 0.375 & 0.379 & 0.291 & 0.426 & 0.601 & 0.290 & 0.367 & 0.435 \\
& 9e-6 & 0.640 & 0.327 & 0.345 & 0.349 & 0.266 & 0.389 & 0.568 & 0.267 & 0.340 & 0.408 \\
& 1e-5 & 0.643 & 0.294 & 0.313 & 0.317 & 0.227 & 0.357 & 0.525 & 0.238 & 0.312 & 0.379 \\
\midrule
\multirow{11}{*}{\makecell{RankPO \\ (Hinge)}}
& 5e-7 & 0.541 & 0.698 & 0.706 & 0.707 & 0.647 & 0.752 & 0.844 & 0.629 & 0.694 & 0.734 \\
& 1e-6 & 0.572 & 0.690 & 0.699 & 0.700 & 0.632 & 0.742 & 0.836 & 0.618 & 0.685 & 0.726 \\
& 2e-6 & 0.578 & 0.680 & 0.690 & 0.691 & 0.619 & 0.730 & 0.831 & 0.605 & 0.672 & 0.716 \\
& 3e-6 & 0.605 & 0.645 & 0.655 & 0.656 & 0.576 & 0.690 & 0.800 & 0.563 & 0.632 & 0.680 \\
& 4e-6 & 0.626 & 0.586 & 0.597 & 0.599 & 0.511 & 0.628 & 0.761 & 0.498 & 0.569 & 0.624 \\
& 5e-6 & 0.644 & 0.508 & 0.522 & 0.524 & 0.431 & 0.560 & 0.709 & 0.422 & 0.496 & 0.556 \\
& 6e-6 & 0.642 & 0.417 & 0.432 & 0.435 & 0.349 & 0.479 & 0.636 & 0.342 & 0.414 & 0.477 \\
& 7e-6 & 0.656 & 0.391 & 0.408 & 0.411 & 0.318 & 0.447 & 0.610 & 0.316 & 0.390 & 0.455 \\
& 8e-6 & 0.644 & 0.345 & 0.363 & 0.367 & 0.273 & 0.400 & 0.574 & 0.278 & 0.351 & 0.418 \\
& 9e-6 & 0.648 & 0.287 & 0.305 & 0.310 & 0.224 & 0.346 & 0.522 & 0.234 & 0.302 & 0.371 \\
& 1e-5 & 0.667 & 0.268 & 0.286 & 0.290 & 0.207 & 0.321 & 0.488 & 0.219 & 0.283 & 0.350 \\
\bottomrule
\end{tabular}
\caption{Performance comparison between RankPO and SFT. 
The table presents the ranking performance of the models on three metrics: MRR, Recall, and nDCG, at three cutoffs $k$ (5, 20, and 100).
The LR column indicates the Learning Rate used, and
the {Align.} column indicates the agreement between the model's ranking and AI preferences. 
RankPO refers to Rank Preference Optimization, while SFT represents Supervised Fine-Tuning. 
RankPO (Sigmoid) refers to RankPO with sigmoid loss, while RankPO (Hinge) denotes RankPO with hinge loss.
Higher values indicate better performance.}
\label{tab:rankpo_performance}
\end{table*}

\begin{table*}[htb]
\centering
\begin{tabular}{llcccccccccc}
\toprule
\multirow{2}{*}{\textbf{Method}} & \multirow{2}{*}{\textbf{LR}} & \multirow{2}{*}{\textbf{Align.}} 
& \multicolumn{3}{c}{\textbf{MRR@k}} 
& \multicolumn{3}{c}{\textbf{Recall@k}} 
& \multicolumn{3}{c}{\textbf{nDCG@k}} \\
\cmidrule(r){4-6} \cmidrule(r){7-9} \cmidrule(r){10-12}
&  &   & \textbf{5} & \textbf{20} & \textbf{100} 
& \textbf{5} & \textbf{20} & \textbf{100} 
& \textbf{5} & \textbf{20} & \textbf{100} \\
\midrule
\multirow{11}{*}{SFT} 
& 5e-7 & 0.563 & 0.692 & 0.701 & 0.702 & 0.639 & 0.739 & 0.833 & 0.623 & 0.687 & 0.729 \\
& 1e-6 & 0.592 & 0.623 & 0.632 & 0.634 & 0.548 & 0.648 & 0.761 & 0.542 & 0.606 & 0.654 \\
& 2e-6 & 0.611 & 0.483 & 0.496 & 0.499 & 0.393 & 0.501 & 0.650 & 0.399 & 0.465 & 0.526 \\
& 3e-6 & 0.609 & 0.377 & 0.392 & 0.395 & 0.289 & 0.390 & 0.545 & 0.304 & 0.367 & 0.431 \\
& 4e-6 & 0.638 & 0.248 & 0.262 & 0.267 & 0.186 & 0.268 & 0.424 & 0.203 & 0.256 & 0.320 \\
& 5e-6 & 0.654 & 0.159 & 0.174 & 0.178 & 0.118 & 0.185 & 0.322 & 0.133 & 0.178 & 0.237 \\
& 6e-6 & 0.672 & 0.107 & 0.119 & 0.123 & 0.074 & 0.134 & 0.249 & 0.090 & 0.129 & 0.182 \\
& 7e-6 & 0.679 & 0.075 & 0.086 & 0.090 & 0.050 & 0.094 & 0.201 & 0.062 & 0.094 & 0.142 \\
& 8e-6 & 0.675 & 0.056 & 0.065 & 0.069 & 0.037 & 0.071 & 0.162 & 0.047 & 0.073 & 0.116 \\
& 9e-6 & 0.671 & 0.047 & 0.054 & 0.057 & 0.031 & 0.059 & 0.136 & 0.040 & 0.062 & 0.099 \\
& 1e-5 & 0.664 & 0.035 & 0.041 & 0.044 & 0.024 & 0.046 & 0.112 & 0.031 & 0.050 & 0.083 \\
\midrule
\multirow{11}{*}{\makecell{SimRankPO \\ (Sigmoid)}}
& 5e-7 & 0.560 & 0.691 & 0.700 & 0.701 & 0.636 & 0.736 & 0.830 & 0.621 & 0.685 & 0.726 \\
& 1e-6 & 0.591 & 0.631 & 0.640 & 0.642 & 0.553 & 0.644 & 0.765 & 0.550 & 0.611 & 0.661 \\
& 2e-6 & 0.600 & 0.524 & 0.536 & 0.539 & 0.435 & 0.544 & 0.684 & 0.438 & 0.504 & 0.562 \\
& 3e-6 & 0.613 & 0.430 & 0.444 & 0.448 & 0.339 & 0.450 & 0.603 & 0.350 & 0.418 & 0.480 \\
& 4e-6 & 0.635 & 0.302 & 0.317 & 0.321 & 0.231 & 0.322 & 0.479 & 0.246 & 0.302 & 0.368 \\
& 5e-6 & 0.650 & 0.219 & 0.233 & 0.238 & 0.165 & 0.244 & 0.390 & 0.183 & 0.233 & 0.296 \\
& 6e-6 & 0.664 & 0.154 & 0.168 & 0.173 & 0.113 & 0.180 & 0.320 & 0.130 & 0.175 & 0.235 \\
& 7e-6 & 0.677 & 0.111 & 0.124 & 0.128 & 0.080 & 0.138 & 0.260 & 0.094 & 0.133 & 0.188 \\
& 8e-6 & 0.671 & 0.090 & 0.102 & 0.107 & 0.064 & 0.114 & 0.225 & 0.076 & 0.111 & 0.163 \\
& 9e-6 & 0.671 & 0.065 & 0.075 & 0.080 & 0.045 & 0.088 & 0.190 & 0.055 & 0.086 & 0.134 \\
& 1e-5 & 0.675 & 0.049 & 0.057 & 0.061 & 0.032 & 0.063 & 0.150 & 0.042 & 0.066 & 0.107 \\
\midrule
\multirow{11}{*}{\makecell{SimRankPO \\ (Hinge)}}
& 5e-7 & 0.555 & 0.693 & 0.703 & 0.704 & 0.641 & 0.742 & 0.835 & 0.624 & 0.690 & 0.730 \\
& 1e-6 & 0.592 & 0.651 & 0.660 & 0.662 & 0.581 & 0.674 & 0.784 & 0.573 & 0.634 & 0.681 \\
& 2e-6 & 0.608 & 0.565 & 0.576 & 0.579 & 0.483 & 0.582 & 0.718 & 0.480 & 0.545 & 0.600 \\
& 3e-6 & 0.611 & 0.489 & 0.502 & 0.505 & 0.400 & 0.509 & 0.654 & 0.403 & 0.471 & 0.531 \\
& 4e-6 & 0.645 & 0.386 & 0.401 & 0.405 & 0.303 & 0.406 & 0.571 & 0.312 & 0.376 & 0.444 \\
& 5e-6 & 0.657 & 0.299 & 0.313 & 0.318 & 0.228 & 0.323 & 0.486 & 0.244 & 0.300 & 0.369 \\
& 6e-6 & 0.666 & 0.211 & 0.226 & 0.230 & 0.155 & 0.235 & 0.380 & 0.174 & 0.225 & 0.287 \\
& 7e-6 & 0.661 & 0.159 & 0.172 & 0.177 & 0.115 & 0.183 & 0.318 & 0.132 & 0.176 & 0.236 \\
& 8e-6 & 0.650 & 0.141 & 0.154 & 0.158 & 0.101 & 0.162 & 0.289 & 0.118 & 0.161 & 0.218 \\
& 9e-6 & 0.681 & 0.104 & 0.116 & 0.121 & 0.075 & 0.127 & 0.240 & 0.090 & 0.127 & 0.179 \\
& 1e-5 & 0.673 & 0.064 & 0.074 & 0.078 & 0.042 & 0.082 & 0.180 & 0.056 & 0.084 & 0.129 \\
\bottomrule
\end{tabular}
\caption{Performance comparison between SimRankPO and SFT. 
The table presents the ranking performance of the models on three metrics: MRR, Recall, and nDCG, at three cutoffs $k$ (5, 20, and 100).
The LR column indicates the Learning Rate used, and
the {Align.} column indicates the agreement between the model's ranking and AI preferences. 
SimRankPO (Sigmoid) refers to SimRankPO with sigmoid loss, while SimRankPO (Hinge) denotes SimRankPO with hinge loss.
Higher values indicate better performance.}
\label{tab:simrankpo_performance}
\end{table*}

\end{document}